\documentclass[runningheads]{llncs}
\usepackage{amsmath}
\usepackage{graphicx}
\begin{document}
\title{Automating Text Naturalness Evaluation of NLG Systems\thanks{Supported by the project no. 19-26934X (NEUREM3) of the Czech Science Foundation and ELITR  (H2020-ICT-2018-2-825460) of the EU.}}  %
\titlerunning{Automating Evaluation of Text Naturalness} 
%
\author{Erion \c{C}ano\orcidID{0000-0002-5496-3860} \and
Ond\v{r}ej Bojar\orcidID{0000-0002-0606-0050}}
\authorrunning{E. \c{C}ano and O. Bojar}
%
\institute{Charles University in Prague, Czech Republic\\
\email{\{cano, bojar\}@ufal.mff.cuni.cz}}

\maketitle              
\begin{abstract}  
Automatic methods and metrics that assess various quality criteria of automatically generated texts are important for developing NLG systems because they produce repeatable results and allow for a fast development cycle. We present here an attempt to automate the evaluation of text naturalness which is a very important characteristic of natural language generation methods. Instead of relying on human participants for scoring or labeling the text samples, we propose to automate the process by using a human likeliness metric we define and a discrimination procedure based on large pretrained language models with their probability distributions. We analyze the text probability fractions and observe how they are influenced by the size of the generative and discriminative models involved in the process. Based on our results, bigger generators and larger pretrained discriminators are more appropriate for a better evaluation of text naturalness. A comprehensive validation procedure with human participants is required as follow up to check how well this automatic evaluation scheme correlates with human judgments.  
\end{abstract}
\section{Introduction}

NLG (Natural Language Generation) is a research direction that develops techniques and practices for automatically transforming structured data into natural language phrases that are readable for humans. Some of the most common applications of NLG are financial reporting, weather predictions, customer support, etc. Same to other related disciplines such as MT (Machine Translation) or ATS (Abstractive Text Summarization), NLG has surged in the last decade, greatly pushed by the significant advances in language applications of deep neural networks \cite{8416973}. In parallel with the data processing, researchers are also exploring possibilities to automate the evaluation of the intelligent text-related systems they propose \cite{DBLP:conf/aaai/00010W018,cano-bojar-2020-two,DBLP:conf/aaai/Zhou020}.
An evaluation practice that is becoming popular recently compares method output samples against human-written references of a standard corpus using automatic metrics. Some of the most popular metrics used for MT and ATS include BLEU of Papineni et al. \cite{P02-1040}, ROUGE of Lin \cite{Lin:2004} and METEOR of Banerjee and Lavie \cite{banerjee-lavie-2005-meteor}. This automatic evaluation practice is appealing for researchers in NLG and NLP (Natural Language Processing) because it is fast and inexpensive to run, does not require domain expertise, and usually yields repeatable and explainable results. 
However, when comparing two methods \textbf{A} and \textbf{B}, we usually want to observe more quality aspects of their produced texts and not just accuracy. These other aspects include, e.g., readability, coherence, naturalness, fluency, adequacy or grammaticality. BLEU and the automatic evaluation process have been criticized by several authors \cite{bojar-etal-2010-tackling,sulem-etal-2018-bleu}. As pointed out by Reiter \cite{reiter-2018-structured}, BLEU is not appropriate for evaluating many quality criteria relevant for NLG systems. Moreover, the results of the automatic evaluation process do not always correlate well with those of human surveys \cite{reiter-belz-2009-investigation}. We would not get any extra information about the text quality aspects of \textbf{A} and \textbf{B} outputs even by adding data efficiency scores in the process \cite{cano-bojar-2019-efficiency}. 
As Novikova et al. \cite{novikova-etal-2017-need} suggests, there is a need for novel evaluation metrics that can objectively assess specific text quality characteristics such as human likeliness (naturalness), readability, coherence, etc.  
In this paper, we present the proof of concept for a method that can potentially be used to automatically evaluate the human likeliness of NLG outputs. This method conceives the text naturalness evaluation from an adversarial perspective, using a discriminator model that can label each test set output as being human-written (natural) or machine-generated (synthetic). This discriminator will use existing pretrained language models such as BERT or GPT2 and presume that synthetic texts do contain relatively more high-rank words sampled from probability distributions learned by the generator, in contrast to natural texts that usually contain more low-rank words. This way, it will compute the fraction of probabilities (the ratio between the probability of the actual word in a position and the highest rank word for that position) for each word and the average value for the entire text. The latter will be discretized to get the class (\emph{h} or \emph{m}) of that text sample.
We start with a review of recent language model applications and observe trends in the ways they are being evaluated. There is actually an increasing trend of automatic evaluations against human-based ones. Later on, in Section~\ref{ssec:hscore}, we propose the \emph{h} score to estimate text naturalness. In Sections~\ref{ssec:hdisc} and \ref{ssec:backup}, the discrimination model is described, together with possible shortcomings and a few alternative approaches. Section~\ref{sec:fracp} shows the results of some ATS experiments we conducted, observing the effect of the generator and discriminator sizes on the probability fractions. Finally, we conclude with the follow-up work of the near future. 

\section{Language Generation Application and Evaluation}
\label{sec:gendisc}

\subsection{Recent Applications of Language Models} %
\label{ssec:genmod}

The NLG progress of the last years has been mostly driven by recent advancements in sequence-2-sequence models based on the encoder-decoder framework. The top sequence matching neural networks have been primarily designed for MT and successfully applied for similar tasks like ATS or NLG \cite{DBLP:journals/corr/BahdanauCB14,P16-1008}. Actually, the numerous NLG applications depend on the ability of each model to learn an appropriate language distribution $p(X_i | X_{1:i-1})$ and use it to decode a word (e.g., the highest probability one) at a time for each position generating phrases, sentences, paragraphs or even longer texts. 
The Transformer architecture introduced by Vaswani et al. \cite{NIPS2017_7181} was a pivotal leap since it improved both performance and training time. It has been widely implemented in several variants, with OpenNMT being the most popular \cite{klein-etal-2017-opennmt}.  
Moreover, Transformer blocks are also used to build big pretrained models such as ELMo (Embeddings from Language Models) of Peters et al. \cite{peters-etal-2018-deep}, BERT (Bidirectional Encoder Representations from Transformers) of Devlin et al. \cite{devlin-etal-2019-bert} or GPT2 (Generative Pretrained Transformer 2) of Radford et al. \cite{noauthororeditor}. These language models are trained on huge amounts of texts and can be tuned with texts of specific tasks, providing state-of-the-art results in MT, ATS, text classification and more.
Because of this research trend, numerous specific BERT variants have been trained and released continuously. Models such as BioBERT (trained with biomedical texts) of Lee et al. \cite{10.1093/bioinformatics/btz682}, SciBERT (trained with scientific texts) of Beltagy et al. \cite{beltagy-etal-2019-scibert}, VideoBERT (trained with video frames and their text descriptions) of Sun et al. \cite{9009570} and similar ones are excelling on their respective tasks. There are also recent studies that have successfully used the original BERT to generate fluent sentences \cite{wang-cho-2019-bert}.
Since the quality of machine-generated texts has been improved dramatically, it is possible today to use big language models for abusive activities such as spreading fake or misleading news, comments or reviews in the social networks or in the Web \cite{fornaciari-poesio-2014-identifying,NIPS2019_9106}. This has created strong incentives for research and development of machine-generated text detection systems \cite{10.1145/3137597.3137600,perez-rosas-etal-2018-automatic}, especially using the adversarial approach with a discriminator that makes use of a language model generator \cite{gehrmann-etal-2019-gltr,NIPS2019_9106}. Other types of learning models are those trained to discriminate between two or more types of language or detect certain aspects in it. Hate (toxic) speech detection models are similar to synthetic text detectors \cite{10.1007/978-3-319-93417-4_48,malmasi:2017:ranlp}. Furthermore, sentiment analysis is an entire research direction that builds models trained with lexical features to sort out positive sentiments from negative ones \cite{nlpinai19,maas-EtAl:2011:ACL-HLT2011,doi:10.1108/DTA-03-2018-0017}. It is interesting to see that the neural networks used to solve these tasks are recently build using language models like BERT \cite{10.1007/978-3-030-36687-2_77}.  

\subsection{Text Detection Systems} %
\label{ssec:detect} 

Several text detection systems proposed recently are being based on pretrained language models. One of them is Grover of Zellers et al. \cite{NIPS2019_9106} which is a left-to-right language model used for fake news generation and detection. It is based on large Transformers and comes with different model sizes, same as GPT2. The model is trained on RealNews, a large news collection they derived from Common Crawl\footnote{\url{https://commoncrawl.org/}} dumps. Authors argue that limiting the generation variance in each decoding step significantly improves the credibility of the synthetic news. They also build a version of their model used as a discriminator and report experiments of paired generator-discriminator models. According to their results, Grover itself is the best detector of its fake news. They also conclude that the size of the discriminator is highly important to get high detection scores.     
QE (Quality Estimation) BERT of Kim et al. \cite{kim-EtAl:2019:WMT} is another case of using a language model like BERT for evaluating natural language generation (more specifically translation). This model is pretrained with parallel data and fine-tuned from QE data. It is then used to assess the translation quality of other models (using HTER score) in the context of  WMT19  QE Shared Task.\footnote{\url{http://www.statmt.org/wmt19/qe-task.html}} Authors report that QE BERT yields significant improvements on both word-level and sentence-level QE tasks. 
Another example is GLTR (Giant Language model Test Room) of Gehrmann et al. \cite{gehrmann-etal-2019-gltr} which can use any language model such as BERT, small GPT2 (117\,M parameters) or large GPT2 (1.5\,B parameters) as backend for detecting synthetic texts of various sizes. Based on their observations, the authors assume that synthetic (machine-generated) texts are mostly created by sampling high-rank words from the head of a language distribution model $p(X_i | X_{1:i-1})$. To assess if a given word is probably sampled from a language distribution, they propose three tests: (i) checking the probability of the given word in relation to the one that was assigned the highest probability $P_{det}(X_i = \hat{X}_i | X_{1:i-1})$; (ii) checking the relative rank of the given word; (iii) checking the entropy of the predicted distribution. Higher values of these scores indicate more chances that the given word and the entire text is synthetic. They also construct a visual tool that highlights text passages and can be used online.\footnote{http://gltr.io/dist/index.html} 

\subsection{Text Quality Evaluation}
\label{sec:nlgeval}

Automatically generated texts should be carefully evaluated from several points of view. Various quality criteria are used on MT and other NLG tasks \cite{denkowski:lavie:amta:2010}. Accuracy is relatively easy to compute in some of the applications where reference texts are available. It usually assesses the lexical similarity between produced text samples and the reference ones utilizing standard metrics such as BLEU or ROUGE. Grammaticality measures the capability of a model to produce grammatically correct texts, with as few mistakes as possible. It is mostly assessed by counting the different types of errors that are found.

Adequacy, on the other hand, rates the amount of meaning that is expressed in a reference sample which is also induced in the respective generated sample. Surveys with human participants and categorical scales are mostly utilized in the process. Clarity and fluency are somehow similar to each other, reflecting how easily is to understand a generated text fragment. They are typically assessed by humans using categorical scales or ranking several alternatives (e.g., produced by alternative models). Finally, human likeliness or naturalness shows the likelihood of a text being natural or written by a human being rather than automatically generated. Besides accuracy, the rest of the above quality criteria are assessed manually by human experts or survey participants.   
To have a quantitative view of the text quality evaluation trends, especially in the context of NLG research, in \cite{DBLP:journals/corr/abs-1902-00753} we examined the papers published in the last five INLG conference proceedings. 
From that survey we noticed an considerable increase in the number of studies (especially those of the last year) using automatic evaluation only. There was also a steady increase in studies using both human and automatic evaluation. Similar facts have been reported by other recent surveys such as the one of Gkatzia and Mahamood \cite{gkatzia-mahamood-2015-snapshot}.   
There was an opposite trend about the works that carry human evaluation only. They have been decreasing steadily, same as reported in \cite{amidei-etal-2018-evaluation} and \cite{8981519}. We also noticed that most human evaluations involve a few domain experts or tens of university students. Furthermore, there is a considerable number of studies that report to have used evaluators crowdsourced from Amazon Mechanical Turk\footnote{\url{https://www.mturk.com/}} or similar platforms. The most frequent text quality criteria they consider are accuracy, readability, coherence, and human likeliness.   

\section{Automating the Human Likeliness Evaluation} 
\label{sec:human}

\subsection{Formalizing the Text Naturalness Score}
\label{ssec:hscore}

From the observations reported in the previous section, it is clear that automating the evaluation of certain text quality criteria such as coherence, fluency or human likeliness is highly desirable.
Some attempts have developed objective measures of text properties like average word length, mean parse tree height, and the average number of nouns \cite{ambati-etal-2016-assessing,Vajjala.Meurers-14-eacl}. These measures are combined in formulas to obtain a score for automatically assessing text readability quality criterion.
In this work, we present a similar attempt regarding the human likeliness or naturalness of machine-generated texts. The \emph{h} score we propose reveals the ability of an NLG model to produce text samples that are human-like or written by humans. This property is substantial and highly desired in NLG applications as well as in various surging computerized tasks such as MT, ATS, question answering and more. 
The traditional approach assesses text naturalness employing human campaigns or surveys that ask participants to rate the texts using point-based schemes. Likert 5-points scale of is one of the most popular methods in the literature \cite{675159b2c1944257ba18910e46cd6dd1,van-der-lee-etal-2019-best}. Our goal is to automate the human likeliness evaluation of any generative model \textbf{G} by considering the task as a binary discrimination problem and computing a metric that we can call the \emph{h} score. Same as the authors of Grover, we considered an adversarial scenario with a generator that produces text samples and a discriminator that tries to find out if they are natural or synthetic using a language model. Let's assume we are using a test set of $n$ reference samples for the evaluation. We can expect to have $n = n_h + n_m$, where $n_h$ is the number of texts that are perceived as human-written and $n_m$ is the number of those which are perceived as machine-generated. Same as in \cite{DBLP:journals/corr/abs-1902-00753}, we consider the \emph{h} score and \emph{m} score of \textbf{G} as the fraction or percentage of its outputs being perceived as human-written or machine-generated. Both metrics can be computed using Equation~\ref{eq:hscore}. 
\begin{equation}  
\label{eq:hscore}  
h^G = \frac{n_h}{n_h + n_m}	\qquad \textup{and} \qquad m^G = 1 - h^G = \frac{n_m}{n_h + n_m} 
\end{equation}
\noindent Instead of relying on human participants for labeling the texts (marking them as class \emph{h} or class \emph{m}) or scoring them (e.g., 1 to 5 as in Likert scale), we suggest automating the task by using a smart discriminator model \textbf{D}. Synthetic texts of today (e.g., the outputs of huge pretrained language models) are very close to those of human professionals. The success of this approach will thus depend on the ability to create a smart discriminator that can recognize such synthetic texts. Using the predictions of \textbf{D} and Equation~\ref{eq:hscore}, we calculate the \emph{h} scores of the methods we wish to evaluate and use it as a model quality indicator. We would normally favor the method which is more capable in fooling the discriminator to think that its text outputs are human-written (resulting in a higher \emph{h} score).       

\subsection{A Binary Discrimination Scheme}
\label{ssec:hdisc}

Our main idea is to adopt the approach of GLTR for constructing the \textbf{D} discriminator of Section~\ref{sec:human}. Pretrained models such as BERT, GPT-2 small or GPT-2 large will be used to get the language distribution $p(X_i | X_{1:i-1})$. Same as in \cite{gehrmann-etal-2019-gltr}, we assume that synthetic texts do mostly contain high-rank terms and natural texts include more low-rank terms. We still need to design a numeric scheme for computing and assessing the quantity of high-rank words used in each sample and a discretization scheme to translate that quantity in one of the two \emph{h} or \emph{m} categories.    
We propose to calculate $Fp$ (fraction of probabilities) for each word of the text sample ($\hat{X}_{1:n}$ word sequence). It is the fraction between the probability of a given word in its position and the highest probability of any language word appearing in that position, given by the language distribution we obtained from the pretrained model. The $Fp$ values of each word will be used to compute the average $Fp$ score of a text sample $s$ consisting of $k$ words using this formula:\vspace{1mm} \begin{center}$Fp_s = 1/k \sum_{i=1}^{k}P(\hat{X}_i)/P(X_i)$\end{center} \vspace{1mm}

\noindent Regarding the discretization, we will need to find a threshold value $Fp_t$ of the $Fp$ score and then separate \emph{h} samples from \emph{m} ones using the scheme of Equation~\ref{eq:classes}:
\begin{equation}
\label{eq:classes}
class(s) =
 \begin{cases}
   h, & \text{if}\ Fp_s < Fp_t \\
   m, & \text{otherwise}
 \end{cases}
\end{equation} 
To find the optimal $Fp_t$ value, we will need to empirically examine many synthetic and natural texts and their respective $Fp$ scores (e.g., following the recommendations of \cite{10.1007/978-3-540-24854-5_85}). 
After labeling every test sample output of the NLG method \textbf{G}, we can make use of Equation~\ref{eq:hscore} to finally obtain its \emph{h} score. A higher \emph{h} score reflects a better ability for producing texts that are perceived as \emph{h} class (texts with more low-rank words). It is thus an indication that the human likeliness of texts produced by \textbf{G} is high.

\subsection{Alternative Schemes} 
\label{ssec:backup}

Since the $Fp$ values are continuous, using a single $Fp_t$ value to separate the categories may not be appropriate. A better alternative could be to use two threshold values for $Fp$: $Fp_l$ as a left boundary and $Fp_r$ as a right boundary. Doing so, we have a better separation of the two intervals for class \emph{h} ($0 < Fp < Fp_l$) and class \emph{m} ($Fp_r < Fp < 1$) by a third interval ($Fp_l < Fp < Fp_r$) that constitutes the \emph{u} (for \emph{undefined} or \emph{unknown}) class of samples. In other words, a better approach could be to use the alternative implementation we proposed at \cite{DBLP:journals/corr/abs-1902-00753} which makes use of the following equations: 
\begin{equation}
\label{eq:classes2}
class(s) =
\begin{cases}
h, & \text{if}\ Fp_s < Fp_l \\
u, & \text{if}\ Fp_l < Fp_s < Fp_r \\
m, & \text{otherwise}
\end{cases}
\end{equation} 
\begin{equation}  
\label{eq:hscore2}
h^G = \frac{n_h}{n_h + n_m + n_u} \qquad \textup{and} \qquad m^G = \frac{n_m}{n_h + n_m + n_u}
\end{equation}
\vskip 2mm
\noindent Once again, to find the optimal $Fp_l$ and $Fp_r$ values, several empirical tests using synthetic and natural samples will need to be conducted. The $Fp_l$ and $Fp_r$ values will be set based on the average natural text $Fp$ and the average synthetic text $Fp$ scores, together with their respective variances.
Another problem could be a information loss from the discretizations schemes of Equations~\ref{eq:classes} or \ref{eq:classes2}. Discretizing the continuous $Fp$ and then computing the continuous \emph{h} score may result in a significant loss of precision. A solution could be to completely avoid the \emph{h} scores and insted use the $Fp$ values of the test samples to compute the average $Fp$ on the entire test dataset. This approach could be cleaner and simpler, leading to a better quality indicator than the \emph{h} score. Furthermore, this simpler practice could be adopted more easily, easing the cross-interpretation of the results.  
Another drawback is the fact that the discrimination scheme could suffer from the word repetitions, grammatically incorrect words, or similar discrepancies. As a result, additional text checkups or preprocessing steps may be required to ensure its validity.    
Finally, the $Fp$ scores depend on the pretrained backend of \textbf{D}. As a result, the human likeliness reports of certain experiments should also include the specific backend used for the discrimination process.

\section{Role of Model Size on Probability Fractions}
\label{sec:fracp}

Since the method and metric we are proposing are completely based on the probability fractions, 
we tried to learn more about the factors that may influence the $Fp$ scores. More specifically, in this section we examine the role of the generator and discriminator sizes in the $Fp$ scores of the texts they produce and assess. 

\subsection{Experimental Setup}
\label{ssec:setup}

The results of predictive models based on neural networks are usually sensitive to the size and depth of the given network. We logically expect the model size (especially the number of Transformer layers) to be important in both generator and discriminator modules. A bigger (deeper) generator does usually yield more coherent and stable texts which are closer to natural ones. Similarly, a bigger and deeper discriminator should be more capable in capturing word and phrase contexts. We thus expect it to be more accurate than a smaller one when setting the relative word ranks and $Fp$ values for each position and the entire text sample. To be more concise, we construct the following two null hypotheses: 
\begin{itemize}
\itemsep0.7em
\item[$\mathbf{H1_0}$] \emph{There is no significant difference between generated and reference sample $Fp$ scores, despite using small and large generators.}
\item[$\mathbf{H2_0}$] \emph{There is no significant difference between generated and reference sample $Fp$ scores, despite using small and large discriminators to compute them.}
\end{itemize}
To confirm or reject these two assumptions, we conducted several experiments with ATS models which read the content of a source text and then learn to generate an abridged abstract of it. We picked the CNN/DailyMail news collection of Nallapati et al. \cite{K16-1028} as our benchmark dataset. It is very popular in the literature since it is one of the few large datasets (287\,113 train, 13\,368 validation, and 11\,490 test samples) with multi-sentence sources and summaries. Our goal is to compare the $Fp$ scores of the generated news summaries against those of the gold or reference ones. For this, we used the predictions and reference samples of CNN/DailyMail test split. 
Since we expect the generated samples to contain relatively more high-rank words, Fp\textsubscript{gen} (average $Fp$ of the generated samples) should be higher than Fp\textsubscript{gold} (average $Fp$ of the reference samples) in most of the cases. Moreover, the difference between Fp\textsubscript{gen} and Fp\textsubscript{gold} should be higher when using a deeper generator compared to a smaller one. Regarding the role of the discriminator size, we also expect to get a higher difference between Fp\textsubscript{gen} and Fp\textsubscript{gold} when using a bigger language model to compute them.  
As the smaller discriminator backend, we used GPT2 (small) which is a stacked decoder Transformer with 12 blocks and a total of 117 million parameters. As the bigger backend, we tried BERT, a bidirectional transformer with 24 blocks and a total of 340 million parameters. We would probably have a more fair comparison by using GPT2 small against GPT2 large. Unfortunately, we could not use the latter because of its size (1.5 billion parameters) and computation requirements.  
The first generator we trained is a text summarizer (here we call it Tsumm) based on a Transformer of four layers in both encoder and decoder, 512 dimensions in each layer, and a total of 81 million training parameters. We used 200K training steps and 8000 warm-up steps, with mini-batches of size 16 and Adam optimizer parameters $\alpha = 0.001, \beta1 = 0.9, \beta2 = 0.999,$ and $\epsilon = 10^{-8}$.
\setlength\tabcolsep{3pt}
\begin{table}[t]
\centering
\caption{\label{tab:samples}Summarization example from Tsumm and Bsumm generators.}
\begin{tabular}{| p{120mm} |}
\hline
\textbf{Story:} this dog 's certainly not setting a good example to the youngsters under her charge . yack:  gemma the pit bull was filmed at home in california being fed some treats . but in a bid to trick her , her owner throws a broccoli spear into the mix . immediately the canine pulls a look of disgust as she chomps on the vegetable . she then proceeds to spit it out on the floor . let 's hope the two children she lives with do n't follow her lead and they learn to love their greens . when she 's not filling her face , other videos show gemma enjoys sleeping and hanging out with her human family . \\
\hline
\textbf{Reference:} gemma the pit bull was filmed at home in california being fed some treats . but in a bid to trick her , her owner throws a broccoli spear into the mix . she then proceeds to spit it out on the floor \\ [0.07ex]
\hline
\textbf{Tsumm:} gemma the pit bull was filmed at home in california being fed some treats . her owner throws a broccoli spear into the mix . she then proceed spit it out on the floor . \\ [0.07ex]
\hline
\textbf{Bsumm:} gemma the pit bull was filmed at home in california being fed some treats . but in a bid to trick her her owner throws a broccoli spear into the mix \\ [0.07ex]
\hline
\end{tabular}
\end{table}
As the bigger generator, we used the abstractive text summarizer of Liu and Lapata \cite{liu-lapata-2019-text}. The authors modify the original BERT to make is suitable for text summarization (encode and manipulate multi-sentence inputs). They overcome the length limit of 512 position embeddings in BERT to process longer source samples. Their BERT variant (they call it BERTSUM) can be used for both extractive and abstractive TS. We used their abstractive summarizer which is an encoder-decoder model with BERTSUM as the encoder and a Transformer of 6 layers as the decoder (we call this Bsumm). To have stability during tuning, the encoder and the decoder are optimized with different learning rates and wormup steps: 0.002, 20000 and 0.1, 10000 respectively. The rest of the parameters are used with the same values as in Tsumm, resulting in a total of about 180 million training parameters. In total, we got four trials: generating with Tsumm and discriminating with GPT2, generating with Tsumm and discriminating with BERT, generating with Bsumm and discriminating with GPT2, and finally, generating with Bsumm and discriminating with BERT. 

\subsection{Probability Fraction Results}
\label{ssec:results}

We ran the experiments and obtained the summaries from the two generators on the CNN/DailyMail test split. Tsumm obtained ROUGE-1, ROUGE-2, and ROUGE-L F$_1$ scores of 38.2\,\%, 16.3\,\%, and 35.3\,\% respectively. The corresponding scores reached by Bsumm were 41.6\,\%, 19.3\,\%, and 38.7\,\%. A story, its reference summary, and the two generated summaries are given in Table~\ref{tab:samples}. As we can see, both Tsumm and Bsumm have produced a coherent and fluent summary (missing only a small portion of the reference one). 
Table~\ref{tab:pfstats} presents the mean and standard deviation $Fp$ values of the predicted and reference news summary samples in each of the four trials.  
We also counted the cases when Fp\textsubscript{gen} is higher than Fp\textsubscript{gold} and computed their difference in each of the four setups. The corresponding results are presented in Table~\ref{tab:pfstats2}. In the first trial (Transumm vs. GPT2), we see that Fp\textsubscript{gen} was higher than Fp\textsubscript{gold} in only 53.76\,\% of the cases. Their difference was only 0.012 and there was no statistical significance. However, the picture seems different in the second trial when BERT is used as the discriminator. Here we have Fp\textsubscript{gen} higher than Fp\textsubscript{gold} in 58.42\,\% of the cases and a score difference of 0.072 which was statistically significant. In the third trial (Bertsumm vs. GPT2) we got 68.32\,\% of the samples having Fp\textsubscript{gen} higher than Fp\textsubscript{gold}. The Fp\textsubscript{gen} - Fp\textsubscript{gold} value was again statistically significant, despite being only 0.051. The last trial gave us even better results, with The Fp\textsubscript{gen} greater than Fp\textsubscript{gold} in 72.64\,\% of the cases and a difference of 0.08 which was again statistically significant. 
\setlength\tabcolsep{5pt}
\begin{table}[t]
\centering
\caption{\label{tab:pfstats}Mean of probability fractions for the two generators and the two discriminators (standard deviation in parenthesis).}
\begin{tabular}{|c|c c|c c|}
\hline
& \multicolumn{4}{|c|}{GPT2\,(117\,M) \qquad\qquad\qquad BERT\,(340\,M)} \\ [0.42ex] 
Generator & Fp\textsubscript{gen} & Fp\textsubscript{gold} & Fp\textsubscript{gen} & Fp\textsubscript{gold} \\ [0.12ex] 
\hline
Tsumm & 0.101\,(0.093) & 0.089\,(0.07) & 0.336\,(0.28) & 0.264\,(0.245) \\
Bsumm & 0.14\,(0.072) & 0.089\,(0.07) & 0.344\,(0.159) & 0.264\,(0.245) \\ 
\hline
\end{tabular}
\end{table}
\setlength\tabcolsep{5pt}
\begin{table}[t]
\centering
\caption{\label{tab:pfstats2}Difference between the mean probability fractions of the generated sample and the gold one. Values with $^\dagger$ are statistically significant ($p < 0.05$).}
\begin{tabular}{|c|c c|c c|}
\hline
& \multicolumn{4}{|c|}{GPT2\,(117\,M) \qquad\qquad\qquad\qquad BERT\,(340\,M)} \\ [0.42ex] 
Generator & Fp\textsubscript{gen} \textgreater~ Fp\textsubscript{gold} & Fp\textsubscript{gen} - Fp\textsubscript{gold} & Fp\textsubscript{gen} \textgreater~ Fp\textsubscript{gold} & Fp\textsubscript{gen} - Fp\textsubscript{gold} \\ [0.12ex] 
\hline
Tsumm & 53.76\,\% & 0.012\,(13.69\,\%) & 58.42\,\% & 0.072$^\dagger$(27.08\,\%) \\
Bsumm & 68.32\,\% & 0.051$^\dagger$(42.79\,\%) & 72.64\,\% & 0.08$^\dagger$(29.46\,\%) \\ 
\hline
\end{tabular}
\end{table}
It is important to note that the $Fp$ values were mostly low which explains the even lower Fp\textsubscript{gen} - Fp\textsubscript{gold} differences (percentage in parenthesis). Based on these results, we can safely reject the two null hypotheses listed above and confirm that larger generators do yield higher differences between $Fp$ scores of their generated samples and the human ones. Furthermore, larger discriminators do also provide higher differences between $Fp$ scores of the synthetic samples and the human ones. These results indicate that the probability fractions could be a viable means for automating text naturalness evaluation. 

\section{Discussion} 
\label{sec:discussion}

In this paper, we started by reviewing several applications of language models to automate various text-related tasks. As our main goal is the automation of text naturalness evaluation, we described a method for automatically evaluating this text quality criterion of the NLG systems by computing the human-likeliness score that we propose. It is the average number of human-labeled test samples. Instead of relying on human participants to label those samples, we propose to use a discrimination approach based on large pretrained language models like BERT or GPT2 and the computation of the probability fraction of each word and the entire text. An alternative scheme can be used to discretize the fraction probability values in cases of high information loss from the discriminator.
To have an idea about the role of the generator and the discriminator in the probability fractions and their viability towards our goal, we conducted several experiments involving two text summarization models and two language models used as discriminators. Our results indicate that the probability fractions produced by larger discriminators can potentially lead to more stable and accurate text naturalness assessments. Furthermore, bigger generation models produce more stable probability fractions. 
In the future, several empirical observations using synthetic and natural samples will be conducted to find the optimal setup of our proposal. We plan to replicate the experiments in similar tasks such as title prediction and question answering. A comprehensive validation of the scheme we proposed by involving human participants who will judge the naturalness of the text samples is also in our schedule. This will check the agreement between automatic predictions and human evaluations.

\bibliographystyle{splncs04}
\bibliography{bib}

\end{document}